\documentclass{article}

\usepackage{amsmath}
\usepackage{amssymb}
\usepackage{mathabx}
\usepackage{color}
\usepackage{multirow}

\usepackage{lipsum}

\usepackage{microtype}
\usepackage{graphicx}
\usepackage{booktabs} 
\usepackage{blindtext, subfig}
\usepackage{dblfloatfix} 

\usepackage{hyperref}

\usepackage[accepted]{icml2020}

\icmltitlerunning{Learning Image Attacks toward Vision Guided Autonomous Vehicles}

\begin{document}

\twocolumn[
\icmltitle{Learning Image Attacks toward Vision Guided Autonomous Vehicles}



\icmlsetsymbol{equal}{*}

\begin{icmlauthorlist}
\icmlauthor{Hyung-Jin Yoon}{unr}
\icmlauthor{Hamidreza Jafarnejadsani}{steven}
\icmlauthor{Petros Voulgaris}{unr}
\end{icmlauthorlist}

\icmlaffiliation{unr}{Mechanical Engineering, University of Nevada, Reno, USA}
\icmlaffiliation{steven}{Mechanical Engineering, Stevens Institute of Technology, USA}

\icmlcorrespondingauthor{Hyung-Jin Yoon}{hyungjiny@unr.edu}

\icmlkeywords{Machine Learning, ICML}

\vskip 0.3in
]



\printAffiliationsAndNotice{}  

\begin{abstract}
    While adversarial neural networks have been shown successful for static image attacks, very few approaches have been developed for attacking online image streams while taking into account the underlying physical dynamics of autonomous vehicles, their mission, and environment. This paper presents an online adversarial machine learning framework that can effectively misguide autonomous vehicles' missions. In the existing image attack methods devised toward autonomous vehicles, optimization steps are repeated for every image frame. This framework removes the need for fully converged optimization at every frame to realize image attacks in real-time. Using reinforcement learning, a generative neural network is trained over a set of image frames to obtain an attack policy that is more robust to dynamic and uncertain environments. A state estimator is introduced for processing image streams to reduce the attack policy's sensitivity to physical variables such as unknown position and velocity. A simulation study is provided to validate the results.
\end{abstract}



\section{Introduction}
We have observed advances in UAV technology that uses machine learning (ML) in computer vision-based applications, e.g., vision-based tracking~\cite{DJI_MAVIC} or navigation. Also, ML-based object detection techniques using camera images have been employed in the perception module to guide autonomous cars~\cite{serban2018standard, Apollo}. Increasing usage of machine learning (ML) tools that use camera input for autonomous vehicles' guidance has brought concerns about whether the ML tools are robust enough against cyber attacks. Recent studies showed that the ML methods are vulnerable to data perturbed by adversaries. To name a few, we list the adversarial attack settings as follows: (1) Adding small perturbations to images enough not to be noticeable to human eyes but strong enough to result in the incorrect image classification~\cite{kurakin2016adversarial_at_scale,kurakin2016adversarial_world,goodfellow2014explaining}; (2) Modifying physical objects such as putting stickers on road~\cite{ackerman2019three} and road sign~\cite{eykholt2018robust}) to fool ML image classifier or end-to-end vision-based autonomous driver. (3) Fooling object tracking algorithm in autonomous driving systems~\cite{jia2020fooling}. 

Depending on the access to the victim model (image classifier or object detector), there are roughly two groups of adversarial attack methods. In the \emph{white-box} attack method, the image attacker generates the adversarial image perturbation through iterative optimization given full access to the victim ML classifier (or object detector)~\cite{kurakin2016adversarial_world, jia2020fooling}. The decision variables in the optimizations are image variables that are high dimensional tensors, and the gradient with respect to the image variables are calculated using backpropagation through the known victim ML classifier~\cite{kurakin2016adversarial_world} (or object detector~\cite{jia2020fooling}). Since assuming full knowledge of model to attack is not realistic, \emph{black-box} approach was introduced~\cite{ilyas2018black, wei2020heuristic}. In the \emph{black box} method, the gradient of the loss function in the optimization is estimated from queried data sample from the victim model (classifier or object detector) instead of directly calculating gradient from the victim model in a setting of Kiefer–Wolfowitz algorithm~\cite{kiefer1952stochastic}. Since the \emph{black-box} approaches~\cite{ilyas2018black} still needs iterative optimization when it needs to generate an adversarial image perturbation, such attack methods are challenging to be implemented into the dynamic system where decisions need to be made given online information.  To overcome the drawback of iterative optimization, the authors in~\cite{xiao2018generating} introduced a generative-neural-network that can skip the iteration steps after the network is trained under the generative adversarial network framework~\cite{goodfellow2014generative}.

\section{Related Work and Proposed Framework}
The attacking perception module of an autonomous driving system was studied~\cite{boloor2020attacking, jia2020fooling,jha2020ml}. In~\cite{boloor2020attacking}, the authors follow up the inspiring demonstration which attacked Tesla's autonomous driving system using stickers on roads~\cite{ackerman2019three} to formulate an optimization problem that finds optimal locations to place black marks on the road which steers end-to-end autonomous driving cars off the road.
In~\cite{jia2020fooling}, the authors demonstrated that the Kalman filter (tracking object) that filters out noisy object detection can still be attacked using a white box method. The authors also demonstrated that attacking the object detection using the Kalman filter (KF) does not necessarily take more images to be perturbed than the object detection without a KF. So, the KF actually did not robustify the autonomous system in the demonstration~\cite{jia2020fooling}. Furthermore, the method in~\cite{jia2020fooling} was shown to be effective in attacking an industry-level perception module that uses vision-based object detection and LIDAR, GPS, and IMU~\cite{Apollo}.  In~\cite{jia2020fooling}, the attack method~\cite{jia2020fooling} uses a neural network to determine the right timing to use the image attack~\cite{jia2020fooling}, given the state of the environment.

However, the adversarial ML attack methods~\cite{jia2020fooling,jha2020ml} toward autonomous systems rely on iterative optimization to generate adversarial image perturbations given each frame of images. A full set of iterative optimization steps are needed for every new image. Therefore, for these methods, it is challenging to respond to changing situations within the control loop run period of the autonomous vehicles. In both papers~\cite{jia2020fooling,jha2020ml}, dealing the issue of iterative optimization for online applications was out of the scope. Furthermore, the attack methods in~\cite{jia2020fooling,jha2020ml} need to use additional state information sometimes less accessible than the image stream, such as the target object being tracked (not only the target class) and the global position of the vehicle.

In this paper, we present an image attack framework for vision-guided autonomous vehicles that learns both an image perturbation generative network~\cite{xiao2018generating} as a low-level attacker, and an attack parameter policy using two time-scales stochastic optimization~\cite{kushner2012stochastic,borkar1997stochastic,heusel2017gans,gupta2019finite} as illustrated in Figure~\ref{fig3_computation_network}. Compared to the previous works~\cite{jia2020fooling,jha2020ml} on attacking autonomous vehicle using adversarial image perturbation, our proposed multi-level framework has the following advantageous features:
\begin{itemize}
    \item A generative network for adversarial image attack is trained to skip the iterative optimization over images as decision variable (which is not suitable for online application).
    \item A state estimator that processes image streams to misguide the vehicle without knowing the victim vehicle's position and velocity state.
\end{itemize}

\section{Learning to Generate Adversarial Images with Multi-level Stochastic Optimization}
\subsection{Problem definition}
Let $\mathcal{X} \subseteq \mathbb{R}^{ W \times H \times C}$ be the image space; Each image has width $W$, height $H$, and color channel $C$. The images are collected from a camera mounted on autonomous vehicles. The images pass through the object detector network $\mathcal{D}: \mathcal{X} \rightarrow \mathcal{Y}$ that maps the images into the detection outcome space $\mathcal{Y}$. The detection outcome is post-processed into the list of bounding box coordinates (center, width, height) of the detected object sorted by the confidence of detection. Denote $(x_t, y_t)$ the image frame from the camera at time $t$ and the detector network's output. Given the estimated target object by the object detector, the autonomous guidance system uses the vehicle's actuators to keep the bounding box of the target at the center of the camera view and the size of the bounding box within a range.  As a result, the vehicle moves toward the target.
\begin{figure}[ht]
\vskip 0in
\begin{center}
\centerline{\includegraphics[width=\columnwidth]{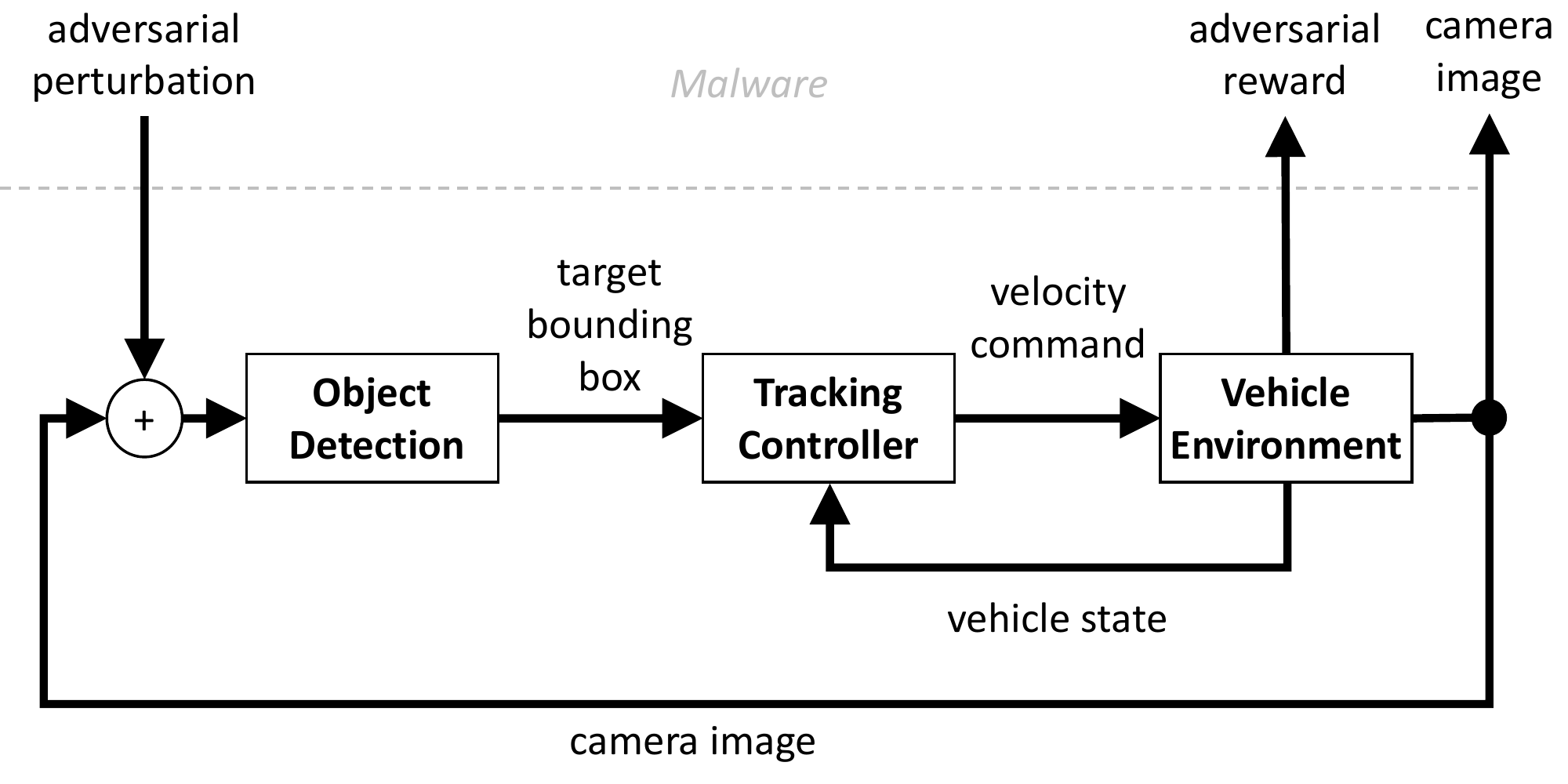}}
\caption{Attacker (malware) and victim system (guidance)}
\label{fig1_victim_sys_malware}
\end{center}
\vskip -0.4in
\end{figure}

As in Figure~\ref{fig1_victim_sys_malware}, it is assumed that the attacker deployed as a \emph{Malware} has access to the image stream and perturbs the image stream input to the object detection module of the victim system. Given the image streams collected until current time $t$ denoted as $\{x_0, x_1, ..., x_t\}$, the goal of the attacker is to generate adversarial image stream $\{\tilde{x}_0, \tilde{x}_1, ..., \tilde{x}_t\}$ which makes the output of the detection network $\{\tilde{y}_0, \tilde{y}_1, ..., \tilde{y}_t\}$ misguide the victim vehicle according to adversarial objectives such as failure of tracking object, moving the vehicle out of the scene, or moving the vehicle to a designated direction. $\tilde{x}_t$ should also be close to the original image $x_t$ in terms of $L_2$ or other distance metric.
\subsection{Proposed framework}
\begin{figure}[ht]
\vskip 0in
\begin{center}
\centerline{\includegraphics[width=\columnwidth]{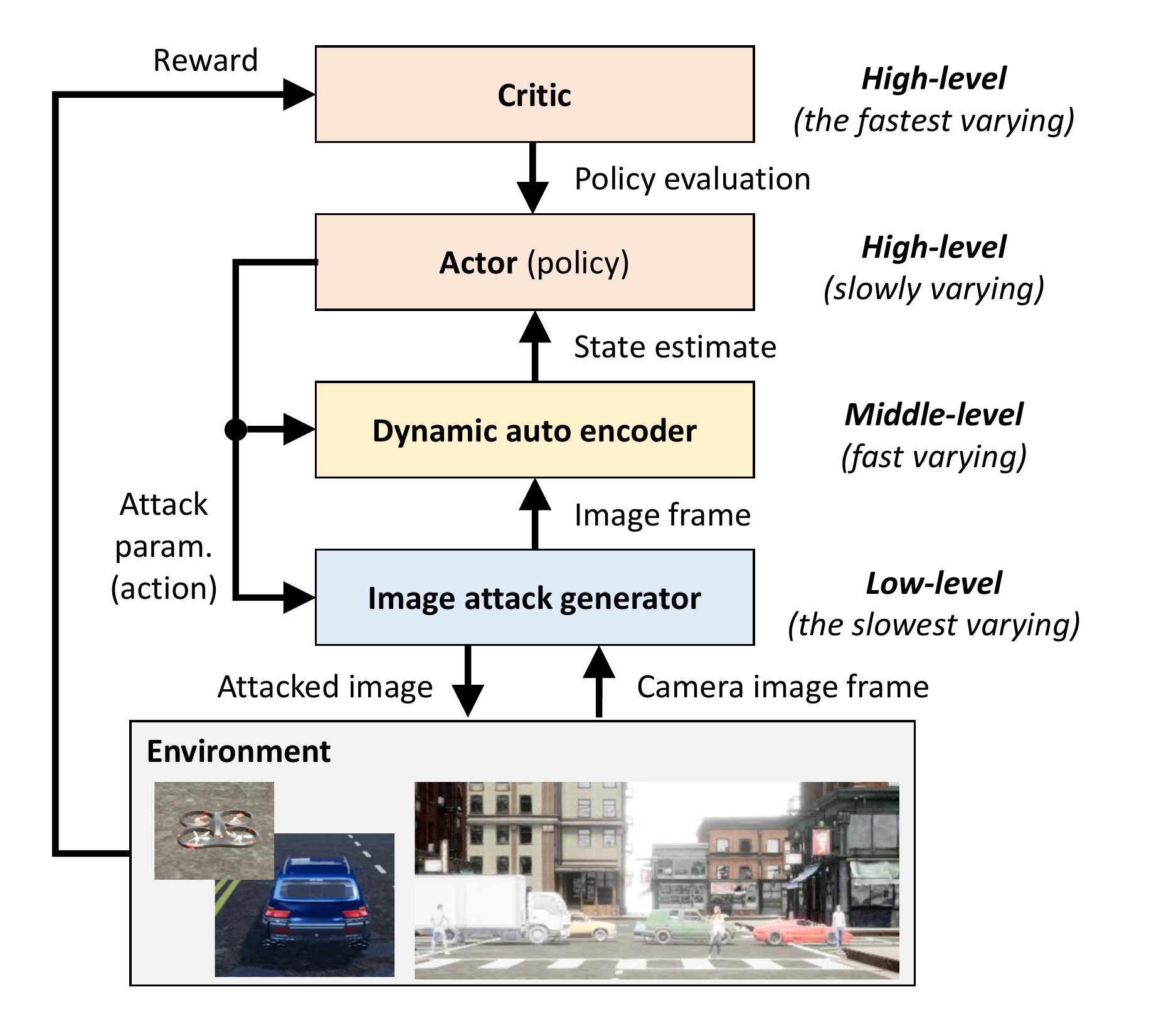}}
\caption{Multi-level image attack framework.}
\label{fig2_proposed_framework}
\end{center}
\vskip -0.4in
\end{figure}
Figure~\ref{fig2_proposed_framework} illustrates the multi-level image attack framework, which consists of four components: critic, actor, dynamic autoencoder, and image attack generator. The image attack generator takes the original image and high-level attack command as its input and generates a perturbation $w_t$. Then $\tilde{x}_t = x_t +  w_t$ is sent to the object detector network $\mathcal{D}$ of the autonomous guidance system (victim).
The perturbed output of the object detector network affects the downstream components in Figure~\ref{fig1_victim_sys_malware} and results in misguiding the victim system. To fulfill the adversarial goal, we first train the image attack generator using the detector network $\mathcal{D}$ as a white-box model. While training the image attack generator, the state estimator (dynamic autoencoder) and the high-level attack are simultaneously trained. We will first describe how the attacker can be used to generate attacks online in~\ref{sec:recursive_attack}, and then the multi-level optimization for training the multi-level framework will be introduced in~\ref{sec:multileveloptim}.

\subsection{Recursive image attack}\label{sec:recursive_attack}
\begin{figure}[h]
\vskip 0.0in
\begin{center}
\centerline{\includegraphics[width=\columnwidth]{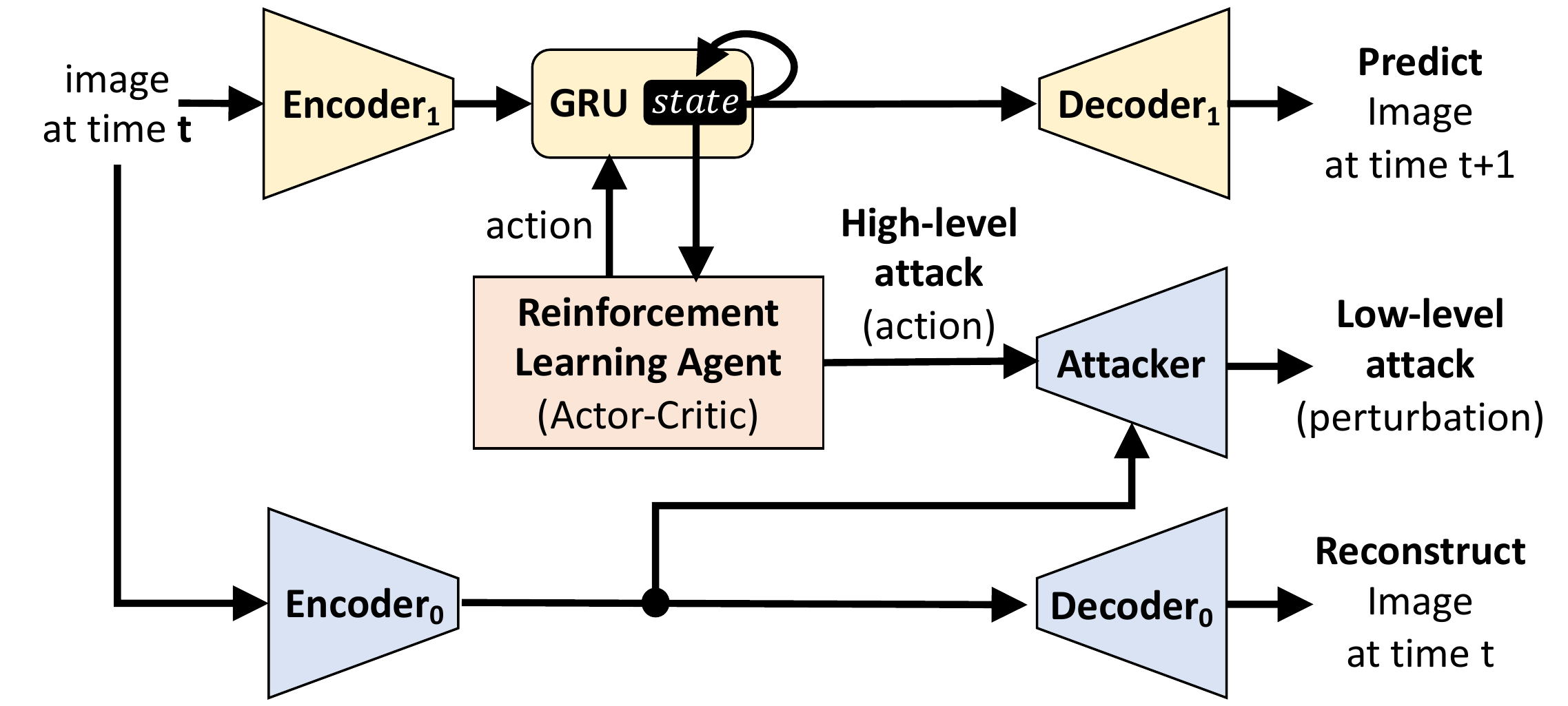}}
\caption{Multi-level image attack computation network. The reinforcement learning agent (the high-level attacker) receives state estimate from dynamic autoencoder and sends high-level attack command to the low-level attacker.}
\label{fig3_computation_network}
\end{center}
\vskip -0.4in
\end{figure}
The computation networks which implement the proposed framework is described in Figure~\ref{fig3_computation_network}. Image $x_t$ at time $t$ is fed into encoder networks for dimension reduction: $\text{Encoder}_1$ for state estimation and $\text{Encoder}_0$ for generating perturbed image $\tilde{x}_t$. The dynamic autoencoder consists of $\text{Encoder}_1$, $\text{GRU}$, and $\text{Decoder}_1$. The $\text{GRU}$ (gated recurrent unit~\cite{cho2014learning}) uses encoded image $\text{Encoder}_1(x_t)$ and the high-level attack action $a_t$ to recursively update the hidden state of GRU, $\mathbf{h}_t$, i.e.,
\begin{equation*}
    \mathbf{h}_{t+1} = \text{GRU}(\mathbf{h}_t, \text{Encoder}_1(x_t), a_t), \quad \mathbf{h}_0 \sim \mathcal{N}(\bar{0},\mathbf{I}).
\end{equation*}
Then $\text{Decoder}_1$ maps $\mathbf{h}_{t+1}$ into a reconstructed image $\hat{y}_{t+1} = \text{Decoder}_1(\mathbf{h}_{t+1})$. We use a distance metric between the image $x_{t+1}$ and the prediction $\hat{y}_{t+1}$ as the loss function to train the dynamic autoencoder. If the dynamic autoencoder can perfectly predict future image $y_{t+1}$ given $\mathbf{h}_t$ and $a_t$ then $\mathbf{h}_t$ has state information consistent with one of the definitions of the state of a system~\cite{bellman1966dynamic}. We need the state estimation since we assume the attacker can only access to the camera image and useful state information such as velocity can not be inferred by independently processing a single image.  

The estimated state information in $\mathbf{h}_t$ is used to generate the high level action by a policy $\mu$ (actor), i.e., $a_t = \mu(\mathbf{h}_t)$ and also for the policy evaluation (critic), i.e., action-value function $Q_{\mu}(\mathbf{h}_t, a_t)$. Then, the adversarial image perturbation is generated by $\text{Attacker}$ given $a_t$ and another encoded image $\text{Encoder}_0$, i.e., $\tilde{x}_t = x_t + \text{Attacker}(\text{Encoder}_0(x_t), a_t)$.

The recursive process of generating adversarial image perturbation using only camera image is summarized in Algorithm~\ref{alg:recursive_attack}.
\begin{algorithm}[ht]
   \caption{Recursive Image Attack}
   \label{alg:recursive_attack}
\begin{algorithmic}
   \STATE {\bfseries Input:} Initial image $x_0$; Initial hidden state of GRU $\mathbf{h}_0$; 
   \STATE {\bfseries Input:} Trained recursive attacker networks;
   \STATE {\bfseries Input:} Autonomous vehicle environment.
   \STATE {\bfseries Output:} Adversarial perturbations $\bar{w}=[w_1, \dots, w_T]$.
   \FOR{$t=0$ {\bfseries to} $T$}
   \STATE Generate attack command using RL policy
   \STATE $\quad a_t \leftarrow \mu(\mathbf{h}_t)$
   \STATE Encode the observed image $x_t$ from the environment
   \STATE $\quad z_t \leftarrow \text{Encoder}_0(x_t)$
   \STATE Generate adversarial image perturbation
   \STATE $\quad w_t \leftarrow \text{Attacker}(z_t, a_t)$
   \STATE Feed $w_t$ to the environment and get new image $x_{t+1}$
   \STATE $\quad x_{t+1}, s_{t+1}, r_{t+1} \leftarrow \text{Environment}(s_t, w_t)$
   \STATE Recursively update the state predictor $\mathbf{h}_{t+1}$ with $x_{t+1}$
   \STATE $\quad \mathbf{h}_{t+1} \leftarrow \text{GRU}(\mathbf{h}_t, \text{Encoder}_1(x_{t+1}))$
   \ENDFOR
\end{algorithmic}
\end{algorithm} 

\subsection{Multi-level optimization to train the attacker}\label{sec:multileveloptim}
We use multi-time scale stochastic optimization to train the multi-level image attack computational networks in Figure~\ref{fig3_computation_network}. Note that the dynamic autoencoder uses the image stream from the environment coupled with the low-level image attacker. Hence, the dynamic autoencoder's adaptation needs to track the environment's changes due to the change of the low-level attacker being updated simultaneously.  Such coupled adaptation issue was considered in multi-level stochastic optimization with hierarchical structures. The multi-level stochastic optimization was used in generative adversarial networks~\cite{heusel2017gans} and reinforcement learning~\cite{konda2000actor}. Employing similar approaches~\cite{heusel2017gans,konda2000actor}, we set the slower parameter update rates in lower-level components.

Let us describe the multi-level stochastic optimization for the parameter iterates: $\theta_n^\text{img}$ of $\text{Encoder}_0(\cdot)$, $\text{Decoder}_0(\cdot)$ and $\text{Attacker}(\cdot)$; $\theta_n^\text{sys}$ of $\text{Encoder}_1(\cdot)$, $\text{Decoder}_1(\cdot)$, and $\text{GRU}(\cdot)$; $\theta_n^\text{actor}$ of the policy  $\mu(\cdot)$; $\theta_n^\text{critic}$ of the critic $Q(\cdot \, , \, \cdot)$ for policy evaluation. The parameters are updated as
\begin{equation}\label{eq:multi-time-scale}
\begin{aligned}
    \theta^\text{img}_{n+1} &= \theta^\text{img}_{n} + \epsilon_n^\text{img} S_n^\text{img}(\mathcal{M}_\text{trajectory})\\
    \theta^\text{actor}_{n+1} &= \theta^\text{actor}_{n} + \epsilon_n^\text{actor} S_n^\text{actor}(\mathcal{M}_\text{transition})\\
    \theta^\text{sys}_{n+1} &= \theta^\text{sys}_{n} + \epsilon_n^\text{sys} S_n^\text{critic}(\mathcal{M}_\text{trajectory})\\
    \theta^\text{critic}_{n+1} &= \theta^\text{critic}_{n} + \epsilon_n^\text{critic} S_n^\text{critic}(\mathcal{M}_\text{transition})
\end{aligned}
\end{equation}
where the update functions $S_n^\text{img}$, $S_n^\text{actor}$, $S_n^\text{sys}$ and $S_n^\text{critic}$ are stochastic gradients with loss functions calculated with data samples from trajectory replay buffer (image stream)
\begin{equation*}
\mathcal{M}_{\text{trajectory}} =
\begin{bmatrix}
(x_{0}, a_{0})_0, &\dots, &(x_t, a_t)_0\\
 \vdots & \vdots & \vdots    \\
(x_{0}, a_{0})_{N}, &\dots, &(x_t, a_t)_{N}
\end{bmatrix}
\end{equation*}
and state transition (previous state, action, reward, state)
\begin{equation*}
\mathcal{M}_{\text{transition}} = 
\begin{bmatrix}
(\mathbf{h}_{t-1-L}, &a_{t-1-L}, &r_{t-1-L}, &\mathbf{h}_{t-L} )\\
 \vdots   &\vdots   &\vdots   &\vdots    \\
(\mathbf{h}_{t-1}, &a_{t-1}, &r_{t-1}, &\mathbf{h}_{t} )
\end{bmatrix}.
\end{equation*}

The step size follows the diminishing rules  
\begin{equation}\label{eq:step-size-rule}
\begin{aligned}
    \epsilon_n^\text{img}/\epsilon_n^\text{sys} \rightarrow 0 \quad
    &\epsilon_n^\text{sys}/\epsilon_n^\text{actor} \rightarrow 0 \quad
    \epsilon_n^\text{actor}/\epsilon_n^\text{critic} \rightarrow 0 \\
    &\text{as} \quad n \rightarrow \infty
\end{aligned}
\end{equation}
as we intended to set slower updates for lower-level components. In~\cite{heusel2017gans}, it was shown that the \emph{ADAM}~\cite{kingma2014adam} step size rule can be set to implement the two time scale step size rule in~\eqref{eq:step-size-rule}. Following the method of using \emph{ADAM} for time-scale separation~\cite{heusel2017gans}, we implemented the step size diminishing rules with \emph{ADAM} optimization rule.   

The multi-time scale stochastic optimization is summarized in Algorithm~\ref{alg:train_attacker}. We describe the loss functions of the stochastic gradients for the multi-level stochastic optimization in the following sections.
\begin{algorithm}
   \caption{Multi-level Stochastic Optimization}
   \label{alg:train_attacker}
\begin{algorithmic}
   \STATE {\bfseries Input:} Recursive Attack Networks in~\ref{fig3_computation_network}
   \STATE {\bfseries \hspace{1cm}} and initial parameters:
   \STATE {\bfseries \hspace{1cm}} $\theta^\text{sys}_0$ of Encoder\textsubscript{0}, GRU, and Decoder\textsubscript{0};
   \STATE {\bfseries \hspace{1cm}} $\theta^\text{actor}_0$ of Policy and $\theta^\text{critic}_0$ of Critic;
   \STATE {\bfseries \hspace{1cm}} $\theta^\text{img}_0$ of Encoder\textsubscript{1}, Decoder\textsubscript{1}, and Attacker.
   \STATE {\bfseries Input:} Autonomous vehicle environment.
   \STATE {\bfseries Input:} Proxy Object Detector $\mathcal{D}$.
   \STATE {\bfseries Input:} Replay buffers: $\mathcal{M}_\text{trajectory}$;$\mathcal{M}_\text{transition}$.
   \STATE {\bfseries Output:} Fixed parameters: $\theta^\text{sys}_*$, $\theta^\text{actor}_*$, $\theta^\text{critic}_*$ and $\theta^\text{img}_*$.
 \STATE {\bfseries Initialize:} $n= n' = 0$
 \REPEAT
 \FOR{$t=0$ {\bfseries to} $T$}
   \STATE Generate and feed the attack into the environment
   \STATE and update state predictor using Algorithm~\ref{alg:recursive_attack}
   \STATE $\quad a_t \leftarrow \mu(\mathbf{h}_t)$
   \STATE $\quad w_t \leftarrow \text{Attacker}(\text{Encoder}(x_t), a_t)$
   \STATE $\quad x_{t+1}, s_{t+1}, r_{t}, \text{done} \leftarrow \text{Env}(s_t, w_t)$
   \STATE $\quad \mathbf{h}_{t+1} \leftarrow \text{GRU}(\mathbf{h}_t, \text{Encoder}_0(x_{t+1}))$
   \STATE Add state transition
   \STATE $\quad \mathcal{M}_\text{transition} \leftarrow (\mathbf{h}_t, a_t, r_{t}, \mathbf{h}_{t+1}) $
   \STATE Update the actor critic parameters: $\theta^\text{actor}_n$, $\theta^\text{critic}_n$
   \STATE with $M$ samples from $\mathcal{M}_\text{transition}$
   \STATE $\quad \theta^\text{critic}_{n+1} \leftarrow \theta^\text{critic}_{n} + \epsilon_n^\text{critic} S_n^\text{critic}(\mathcal{M}_\text{transition})$
   \STATE $\quad \theta^\text{actor}_{n+1} \leftarrow \theta^\text{actor}_{n} + \epsilon_n^\text{actor} S_n^\text{actor}(\mathcal{M}_\text{transition})$
   \STATE where $S_n^\text{critic}$ and $S_n^\text{actor}$ are defined in~\eqref{eq:critic} and~\eqref{eq:grad_actor}.
   \STATE $n \leftarrow n+1$ 
   \IF{\text{done}}
   \STATE $T_\text{stop} \leftarrow t$ and \textbf{break}
   \ENDIF
 \ENDFOR
   \STATE Add a trajectory
   \STATE $\quad \mathcal{M}_\text{trajectory} \leftarrow ((y_0,a_0) \dots, (y_t,a_t)) $
 \FOR{$t=0$ {\bfseries to} $T_\text{stop}$}
   \STATE Update the parameters: $\theta^\text{sys}_{n'}$, $\theta^\text{img}_{n'}$
   \STATE with $M$ samples from $\mathcal{M}_\text{tran}$
   \STATE $\quad \theta^\text{sys}_{n'+1} \leftarrow \theta^\text{sys}_{n'} + \epsilon_n^\text{sys} S_n^\text{critic}(\mathcal{M}_\text{trajectory})$
   \STATE $\quad \theta^\text{img}_{n'+1} \leftarrow \theta^\text{img}_{n'} + \epsilon_n^\text{img} S_n^\text{img}(\mathcal{M}_\text{trajectory})$
   \STATE where $S_{n'}^\text{sys}$ and $S_{n'}^\text{img}$ are defined in~\eqref{eq:grad_sys_id} and~\eqref{eq:grad_img_attack}.
   \STATE $n' \leftarrow n'+1$ 
 \ENDFOR
 \UNTIL{Rewards $r_t$ meets the requirements.}
\STATE \textbf{Fix} the output parameters with the current ones.
\end{algorithmic}
\end{algorithm}

\subsubsection{Image attack generator}\label{sec:image_attack_generator}
We use a proxy object detector (white box model) to train the attack generator. The proxy object detector is \emph{YOLO9000}~\cite{redmon2017yolo9000} pre-trained with \emph{COCO} dataset~\cite{lin2014microsoft}. The detector network $\mathcal{D}$ maps an image $x_t$ into a tensor, which contains bounding box predictors for each anchor points: 
\begin{equation*}
    y_t = \mathcal{D}(x_t), \quad y_t\in\mathbb{R}^{B_\text{yolo} \times W_\text{yolo} \times H_\text{yolo} \times D_\text{yolo}}
\end{equation*}
where $B_\text{yolo}=5$ is the number of the bounding boxes for each anchor point, $W_\text{yolo}=14$, $H_\text{yolo}=14$ denote the number of anchor points along with horizontal and vertical directions on the rectangular image, and $D_\text{yolo}$ denotes the number of features associated with the corresponding bounding box at the anchor point. The features include the coordinate of the bounding box and object classification probabilities~\cite{redmon2017yolo9000}. 

In~\cite{jia2020fooling}, the image attacker moves the center coordinate of a target bounding box used by the victim system's guidance system. It needs more accessibility for the attacker to know which object is being tracked. However, we do not assume that the target bounding box is given to the attacker. Instead, the high-level attacker chooses an anchor point to detect non-existing objects at the anchor, to override the current target object.

Given high level attack $a_t \in [0,1]\times[0,1]\times[0,1]$, we first discretize it into the index of the target anchor, i.e., $\tilde{a}_t =\text{discrete}(a_t) \in I \times J \times K$ where $I=\{1,\dots B_\text{yolo}\}$, $J=\{1,\dots, W_\text{yolo}\}$ and $K=\{1,\dots, H_\text{yolo}\}$. Training of the image attack aims to minimize the following loss function, given the target index $\tilde{a}_t=(i,j,k)$,
\begin{equation}\label{eq:loss_img_attack}
\begin{aligned}
    & l^{img}(x_t, a_t;\theta^\text{img}) \\
    &=  h\left([\mathbf{P}(\mathcal{D}(\tilde{x}_t))]_{i,j,k}, 1\right) \\
    &+ \lambda_0 H\left(\mathbf{P}(\mathcal{D}(\tilde{x}_t)), \bar{\mathbf{0}}\right) \\
    &+ \lambda_1 H\left(x_t, \text{Decoder}_0(\text{Encoder}_0(x_t))\right) \\
    &+ \lambda_2 d(x_t, \tilde{x}_t)
\end{aligned}
\end{equation}
where $\tilde{x}_t$ denotes the attacked image and let us describes each terms in the loss function. In the first term $h\left([\mathbf{P}(\mathcal{D}(\tilde{x}_t))]_{i,j,k}, 1\right)$, $\mathbf{P}(\mathcal{D}(\tilde{x}_t))$ denotes the detector's predicted probability of having the target class at each anchor point, i.e., $[\mathbf{P}(\tilde{y}_t)]_{i',j',k'} = \text{\emph{Likelihood}}\{\text{The anchor\textsubscript{i',j',k'} detects the target class.}\}$. Hence, the binary cross entropy\footnote{For $y\in [0,1]$ and $\hat{y}\in (0,1)$, the binary cross entropy is calculated as $h(y, \hat{y})=y\log\hat{y} + (1-y)\log(1-\hat{y})$ and we follow the convention $0 = 0 \log 0$.} $h(\cdot, \cdot)$ in the second term is to promote the target class is detected at the anchor $(i,j,k)$ by the detector.

In the second term $H\left(\mathbf{P}(\mathcal{D}(\tilde{x}_t)), \bar{\mathbf{0}}\right)$, $\bar{\mathbf{0}}$ denotes a zero tensor with the same shape of $\mathbf{P}(\mathcal{D}(\tilde{x}_t))$ and $H\left(\mathbf{P}(\mathcal{D}(\tilde{x}_t)), \bar{\mathbf{0}}\right)$ is calculated as
\begin{equation*}
    H(\mathbf{P}, \bar{\mathbf{0}}) = \frac{1}{B_\text{yolo}W_\text{yolo}H_\text{yolo}}\sum_{I,J,K}h([\mathbf{P}]_{i',j',k'}, 0).
\end{equation*}
Therefore, the second term is to vanish the detected bounding boxes of the target class. 

The third term is to train the autoencoder reconstruction loss to compress the image for attack image generation, i.e., $H(x_t, \hat{x}_t) = \frac{1}{ W H C}\sum_{i=1}^W  \sum_{j=1}^H \sum_{k=1}^C h([x]_{i,j,k},[\hat{x}]_{i,j,k})$. And the last term $d(x_t, \tilde{x}_t)$ denotes a distance metric between the attacked image and the original image. $\lambda_0$, $\lambda_1$ and $\lambda_2$ are weighting parameters of the loss function in~\eqref{eq:loss_img_attack}.

In addition to the distance metric $d(x_t, \tilde{x}_t)$ to constrain the perturbation, the adversarial image perturbation $w_t \in [0,1]^{W \times H \times C}$ is added with a hard constraint for original image with color channel signal values in $[0,255]$ as  
\begin{equation*}
    \tilde{x}_t= \max(\min(x_t + \alpha \, w_t, 255),0)
\end{equation*}
where $w_t = \text{Attacker}(\text{Encoder}_0(x_t))$ and $\alpha$ is attack size level which was set to $10$ or $20$ in the numerical examples of this paper.

With the loss function in~\eqref{eq:loss_img_attack} and mini-batch data $(x_1, a_1, \dots, x_M, a_M)$ uniformly and independently sampled from $\mathcal{M}_\text{trajectory}$, we define the stochastic gradient for training image attacker as
\begin{equation}\label{eq:grad_img_attack}
\begin{aligned}
    &S^\text{img}_n(\mathcal{M}_\text{trajectory};\theta_\text{img}) \\
    &=  - \nabla_{\theta_\text{img}} \left( \frac{1}{M} \sum_{m=1}^M l^\text{img}(x_m, a_m;\theta_\text{img}) \right).
\end{aligned}
\end{equation}

\begin{figure}
\vskip 0.0in
\begin{center}
\centerline{\includegraphics[width=\columnwidth]{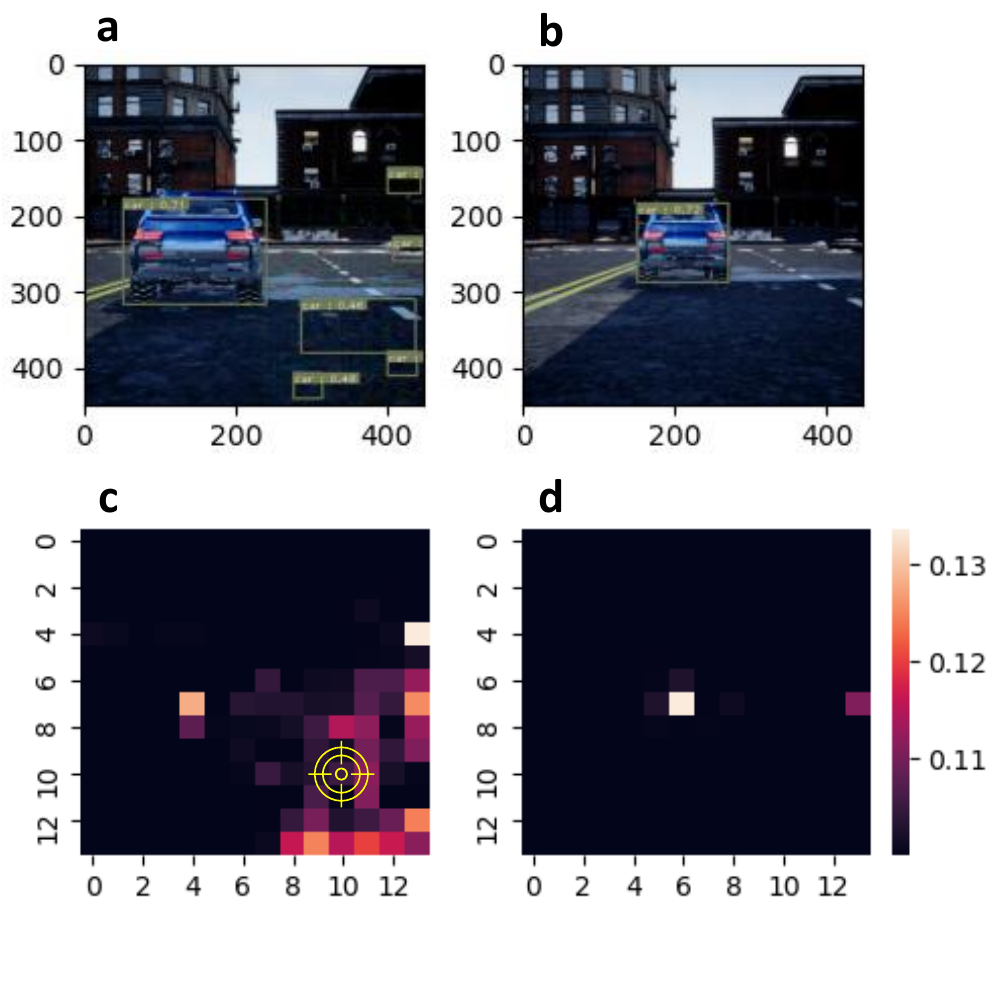}}
\vskip -0.4in
\caption{Attacked image and detection of cars:(a) Attacked image $\tilde{x}_t$ and bounding boxes. (b) Original image $x_t$ and detected bounding box. (c) Heat map (averaged along $I$) of predicted probability $\mathbf{P}(\mathcal{D}(\tilde{x}_t))$  with the attacked image at the anchor of $(10,10,3)$. (d)  Heat map of predicted probability $\mathbf{P}(\mathcal{D}(x_t))$ in~\eqref{eq:loss_img_attack}.}
\label{figure4_lowlevelattack}
\vskip -0.2in
\end{center}
\end{figure}

The low-level attack is illustrated in Figure~\ref{figure4_lowlevelattack}, which shows how the high-level attack $a_t$ steers the output of the detector network and the detected bounding box.  

\subsubsection{System identification for state estimation}
The system identification aims to determine the parameter which maximizes the likelihood of the state estimate. We maximize the likelihood of state predictor by minimizing the cross-entropy error between true image streams and the predicted image streams by a stochastic optimization which samples trajectories saved in memory buffer denoted $\mathcal{M}_\text{trajectory}$ with a loss function to minimize.

The loss function $l^\text{sys}(\cdot)$ is calculated with sampled trajectories from $\mathcal{M}_{\text{trajectory}}$ as
\begin{equation*}
\begin{bmatrix}
(x_{t_0^1}, a_{t_0^1}, & \dots &x_{t_T^1}, a_{t_T^1})_1\\
 \vdots & \vdots &  \vdots\\
(x_{t_0^M}, a_{t_0^M}, &\dots &x_{t_T^M}, a_{t_T^M})_M
\end{bmatrix}
\end{equation*}
where ${t_0^m, t_1^m, \dots, t_T^m}$ denote the time indice with the random initial time and the length $T$ of $m$\textsuperscript{th} sampled trajectory.
We calculate the loss function as
\begin{equation}\label{eq:loss_sys_id}
    l^\text{sys}(\mathcal{D}_n^\text{trajectory};\theta_\text{sys}) = \frac{1}{M}\sum_{m=1}^M H(\mathbf{X}_m, \hat{\mathbf{X}}_m)
\end{equation}
where $\mathbf{X}_m = (x_0, \dots, x_T)_m$ is the $m$\textsuperscript{th} sample image stream with time length $T$. Here, $H(\cdot, \cdot)$ is average of the binary cross-entropy $h(\cdot, \cdot)$ between the original image stream $\mathbf{X}_m$ and the predicted image stream $\hat{\mathbf{X}}_m$ as
\begin{equation*}
    H(\mathbf{X}_m, \hat{\mathbf{X}}_m) = \sum_{t=1}^{T}\sum_{i=1}^{W}\sum_{j=1}^{H}\sum_{k=1}^{C} h([\mathbf{X}_m]_{t,i,j,k}, [\hat{\mathbf{X}}_m]_{t,i,j,k})
\end{equation*}
where $i,j,k$ denotes width, height, color index for the image with width $W$, height $H$, color channel $C=3$ and the cross entropy $h(\cdot, \cdot)$ calculate the difference between the pixel intensities of $[\mathbf{X}_m]_{t,i,j,k} \in [0,1]$ and   $[\hat{\mathbf{X}}_m]_{t,i,j,k} \in (0,1]$.

We generate the predicted trajectory given the original trajectory with image stream $\mathbf{X}_m = (x_0, \dots, x_T)_m$ and action stream $(a_0, \dots, a_T)_m$ by processing them through the encoder, GRU, and the decoder as
\begin{equation*}
    \begin{aligned}
    \mathbf{h}_{t+1} &= \text{GRU}(\mathbf{h}_t, \text{Encoder}_1(y_t), a_t), \quad \mathbf{h}_0 \sim \mathcal{N}(0,\mathbf{I})\\
    \hat{x}_{t+1} & = \text{Decoder}_1(\mathbf{h}_{t+1})
    \end{aligned}
\end{equation*}
and collect them into $\hat{\mathbf{X}}_m =\{\hat{x}_1, \dots, \hat{x}_T\}$.

With the loss function in~\eqref{eq:loss_sys_id}, the stochastic gradient for the optimization is defined as
\begin{equation}\label{eq:grad_sys_id}
    S^\text{sys}_n =  -\nabla_{\theta_\text{sys}} l^\text{sys}(\mathcal{M}_\text{trajectory};\theta_\text{sys}).
\end{equation}

\subsubsection{Actor-Critic Policy Improvement}
The critic evaluates the policy relying on the optimally principle~\cite{bellman1966dynamic} and the use of the principle is helpful to control the variance of the policy gradient by invoking \emph{control variate} method~\cite{law2000simulation}. We employed an actor-critic method~\cite{lillicrap2015continuous} for the reinforcement learning component in the proposed framework. 

The critic network calculates the future expected rewards given the current state and the action, i.e., $\mathbb{E}\left[\sum \gamma^t r_t| \mathbf{h}_0 = \mathbf{h}, a_0 = a\right] \approx Q(\mathbf{h}, a)$. The critic network is updated with the stochastic gradient as 
\begin{equation}\label{eq:critic}
    S^\text{critic}_n = -\nabla_{\theta_\text{critic}} l^\text{critic}(\mathcal{M}_\text{transition};\theta_\text{critic})
\end{equation}
with the following loss function\footnote{We follows the approach of using target networks in~\cite{lillicrap2015continuous} to reduce variance during stochastic updates.}
\begin{equation}\label{eq:loss_critic}
\begin{aligned}
    &l(\mathcal{D}_n^\text{transition};\theta^\text{critic})\\ &=\frac{1}{M}\sum_{m=1}^M(Q(\mathbf{h}_m, \mathbf{a}_m; \theta^\text{critic})- Q_m^{\text{target}})^2, \\
\end{aligned}
\end{equation}
where $Q_m^{\text{target}} = r_m + \gamma Q(\mathbf{h}'_{m}, \mu_\theta(\mathbf{h}'_m);\theta_\text{critic})$ and the state transition sample $[(h, a, h', r)_0, \dots, (h, a, h', r)_M]$ are sampled from the following replay buffer $\mathcal{M}_{\text{transition}}$.

With the same state transition data samples, we calculate the stochastic gradient for the policy as
\begin{equation}\label{eq:grad_actor}
    S^\text{actor}_n = \nabla_{\theta_\text{actor}} J(\mathcal{M}_\text{transition};\theta_\text{actor})
\end{equation}
using the estimated policy value
\begin{equation}\label{eq:loss_actor}
    J = \frac{1}{M}\sum_{m=1}^M Q(\mathbf{h}_m, \mu(\mathbf{h}_m; \theta_\text{actor});\theta_\text{critic})
\end{equation}

\section{Experiments}
We present experiments with our attackers to evaluate the ability to misguide the vehicles in three simulation scenarios: (1) moving a UAV away from the scene where the target object is in. (2) driving a UAV to a direction set by the attacker. (3) losing the tracking of a Kalman filter object tracking for a car follower.

\subsection{Experimental setup}
All our experiments consider attacking a vision based guidance system depicted in Figure~\ref{fig1_victim_sys_malware} that uses YOLO object detector~\cite{redmon2017yolo9000} trained with \emph{COCO} dataset~\cite{lin2014microsoft}. The vehicle simulation environment is implemented with \emph{AirSim}~\cite{airsim2017fsr} and \emph{Unreal} game-engine~\cite{Unreal}. The control loop frequency of receiving a camera image, updating the state estimator, generating an attack, and feeding in the attacked image is $10$ Hz. Episodic training is used to optimize the computational networks in Figure~\ref{fig3_computation_network} as in deep reinforcement learning~\cite{lillicrap2015continuous}. The implementation of Algorithm~\ref{alg:recursive_attack} and Algorithm~\ref{alg:train_attacker} and the experiment environments can be found in the code repository uploaded in supplementary materials.

\subsection{Scenarios and attack performance}
\begin{figure*}[ht]
\centering
\subfloat[][Move away. \href{https://youtu.be/bsQUdH400fI}{video link}]{\includegraphics[width=0.32\linewidth]{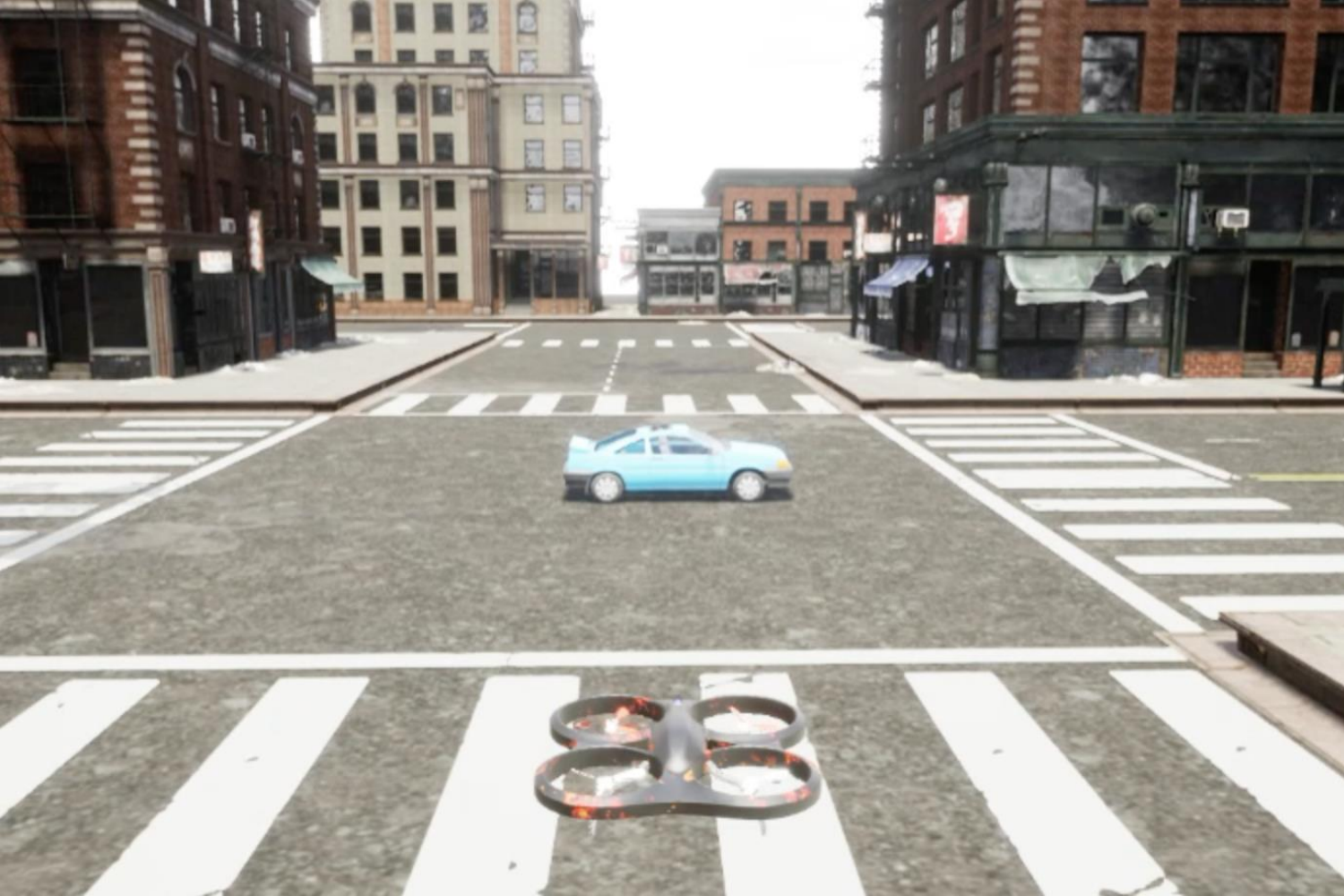}
}
\subfloat[][Move to the right. \href{https://youtu.be/FvyvWEMjiy0}{video link}]{\includegraphics[width=0.32\linewidth]{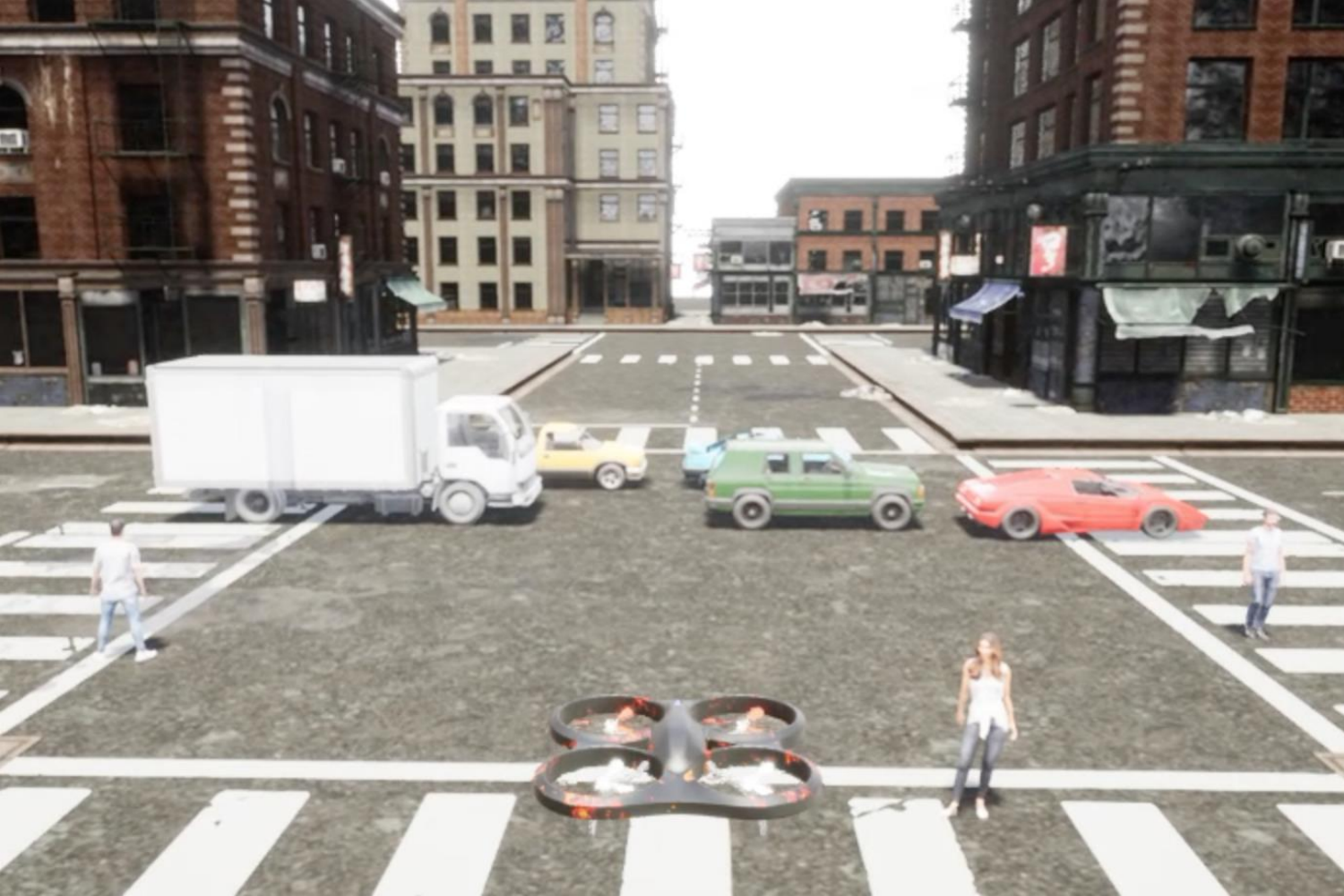}
}
\subfloat[][Lost tracking. \href{https://youtu.be/BZydqH1fNEc}{video link}]{\includegraphics[width=0.32\linewidth]{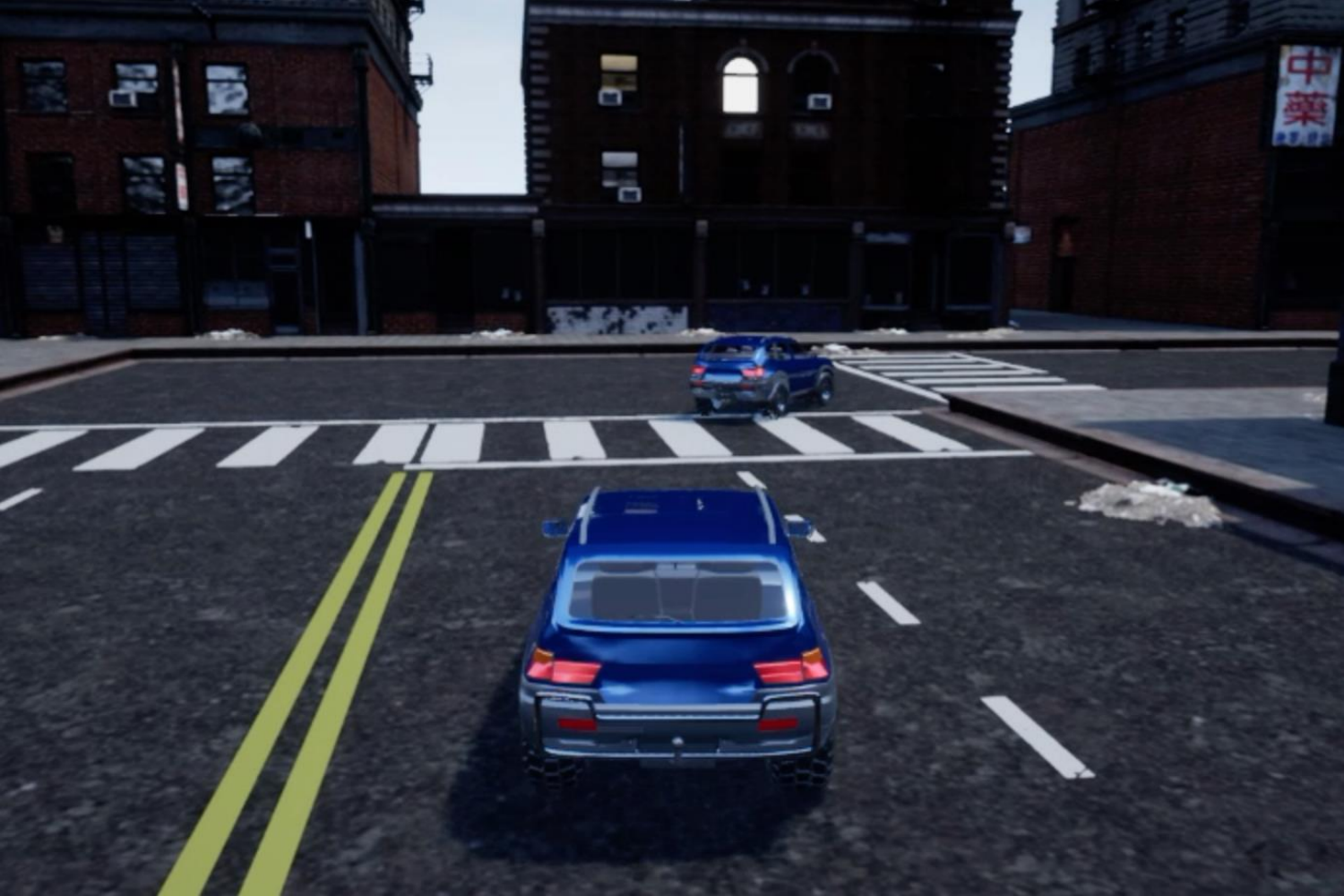}
}
\caption{Simulation scenarios: (a) Moving an UAV away. (b) Moving an UAV to a direction. (c) Attacking object tracking.}
\label{fig:environemnts}
\end{figure*}
\begin{figure*}[ht]
\centering
\subfloat[][Move away. ]{\includegraphics[width=0.32\linewidth]{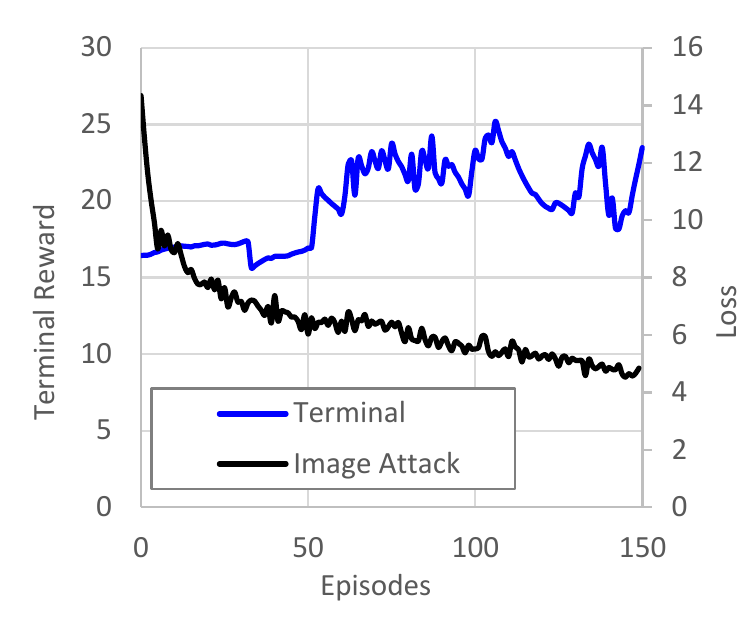}
 }
\subfloat[][Move to the right.]{\includegraphics[width=0.32\linewidth]{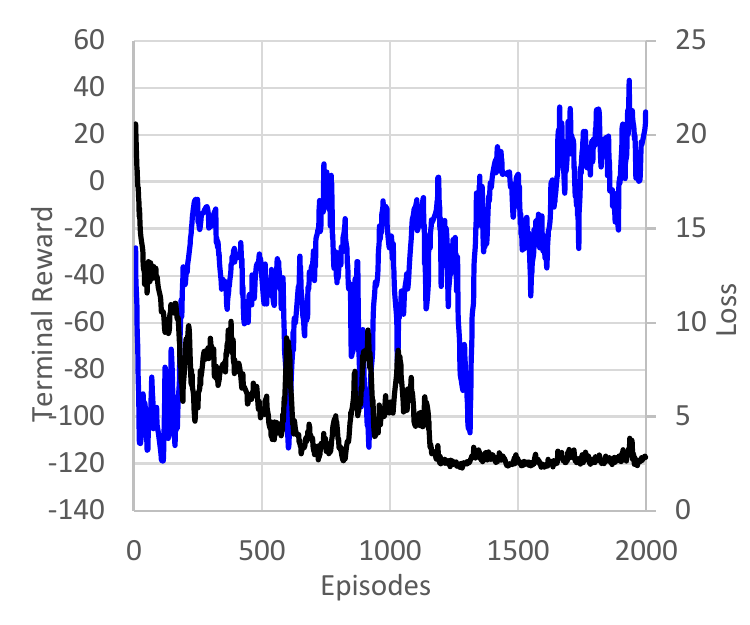}
 }
\subfloat[][Lost tracking.]{\includegraphics[width=0.32\linewidth]{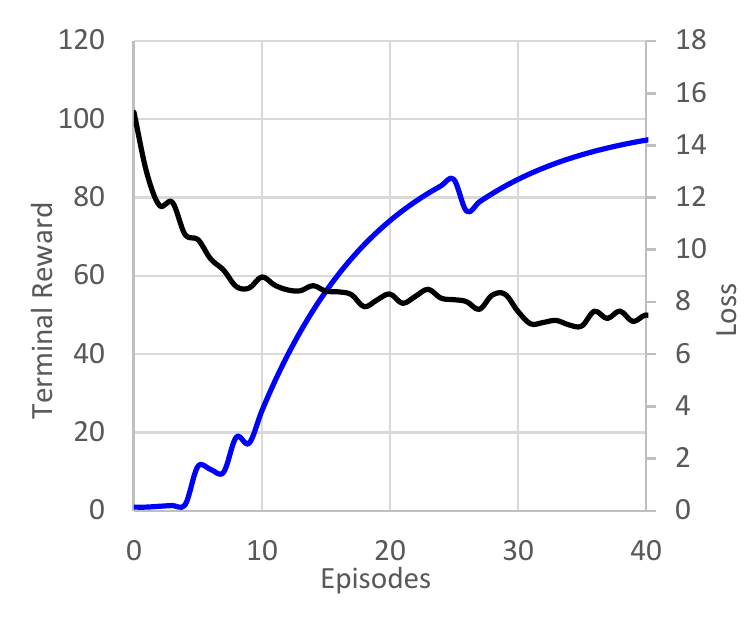}
}
\caption{Reinforcement learning reward iterates and low-level image attacker loss iterates.}
\label{fig:iterations}
\end{figure*}

\subsubsection{Moving an UAV away from the scene}\label{sec:secenario1}
The first scenario of image attack is on a UAV moving toward a car at an intersection, as shown in Figure~\ref{fig:environemnts}a. As in Figure~\ref{fig1_victim_sys_malware}, the guidance system (victim) first detects the car in the scene as a bounding box and calculates the relative position error from the UAV to the target for generating velocity commands (e.g., proportional derivative control). The lateral and vertical position error is calculated in terms of pixel counts on the image frame grid from the center of the camera view to the target bounding box center. The longitudinal position error is calculated from the area of the bounding box. The tracking controller in Figure~\ref{fig1_victim_sys_malware} aims to keep the target bounding box at the center for the camera view and the area of the bounding box within a designated range to keep distance between the UAV and the target at a target distance.

In this scenario, the attacker's goal is to hinder the tracking controller and eventually move the UAV away from the target. The attacker is trained using the multi-level stochastic  Algorithm~\ref{alg:train_attacker} given the information of the victim systems: (1) The proxy object detector (the copy of the same object detector used by the guidance system); (2) target class, which is a car here; (3) camera image stream and the reward which is devised according to the attacker's objective. The running reward function is defined as
\begin{equation*}
    r(s_t, a_t) = 
    \begin{cases}
        0.1 & \text{if} \quad v_\text{uav} > 0.1 \text{ [m/s]}\\
        -0.1 & \text{otherwise.}
    \end{cases}
\end{equation*}
Each episode terminates when the UAV stops moving, and the terminal reward is the distance from the origin. The terminal reward (performance metric) is the attack performance metric for evaluations compared to normal episodes (without attack). 

As shown in Figure~\ref{fig:iterations}a, through the 150 episodes of training, the terminal reward has increased while the low-level image attack training loss was decreasing.
As in the first row of Table~\ref{tab:reward-metric}, the attacker metric is greater\footnote{We used the Mann Whitney U test to determine whether the attacker makes statically significant differences (significance level at $0.01$).} when we use the attacker than in the normal episodes.

\subsubsection{Moving an UAV to a direction}
In this scenario, the same victim system in~\ref{sec:secenario1} is used in a similar environment where more cars and three people on the street as shown in Figure~\ref{fig:iterations}b. The guidance system brings the UAV to the most car-like object (i.e., with the greatest confidence in the object detector's output). 

The attacker's goal in this scenario is to steer the UAV to the right. For the objective, we define the running reward as
\begin{equation*}
    r(s_t, a_t) = 
    \begin{cases}
        0.1 & \text{if} \quad \mathbf{v}_y > 0.01 \text{ [m/s]}\\
        -0.1 & \text{otherwise}
    \end{cases}
\end{equation*}
Each episode terminates when the UAV stops moving, and the terminal reward (performance metric) is proportional to the distance moved to the right.

As shown in Figure~\ref{fig:iterations}b, through the 2000 episodes of training, the terminal reward has increased while the low-level image attack training loss was decreasing. As in the second row of Table~\ref{tab:reward-metric}, the attacker metric is greater when we use the attacker than in the normal situation (without attack).

\subsubsection{Attacking Kalman filter object tracking}
In this scenario in Figure~\ref{fig:environemnts}c, we are using a car chasing environment where the guidance system does not use the bounding box directly but a filtered bounding box using a double integrator model with a Kalman filter (KF) as in~\cite{jia2020fooling}.

The attacker's goal in this scenario is to lose the following car's tracking of the front car. For this objective, we define the running reward as proportional to the distance between the car. The episode terminates if the second car collides or the time passed the time limit with the following terminal reward (performance metric) as
\begin{equation*}
    r_\text{terminal} = 
    \begin{cases}
        100 & \text{if the car collides}\\
        0   & \text{if $t > 500$ steps}
    \end{cases}
\end{equation*}

As shown in Figure~\ref{fig:iterations}c, through the 40 episodes of training, the terminal reward has increased while the low-level image attack training loss was decreasing. As in the third row of Table~\ref{tab:reward-metric}, the attacker metric is greater when we use the attacker than in the normal situation.

\begin{table}[ht]
\caption{Attack metric and comparison statistics of normal episodes and attack episode (10 runs each).}
\label{tab:reward-metric}
\begin{center}
\begin{small}
\begin{sc}
\begin{tabular}{cccc}
\toprule
                  &          & Attack            & Mann\\
                  &          & Metric            & Whitney \\
                  &          & (avg. $\pm$std.)  & U test \\
\midrule
Scenario 1        & Normal   & $16.7\pm0.16$     &  $100.0$   \\  
\emph{away}       & Attack   & $33.3\pm2.10$     &  ($p < 0.01$)      \\ 
\midrule
Scenario 2        & Normal   & $-1.36\pm0.415$   & $87.0$     \\
\emph{to right}   & Attack   & $0.048\pm0.951$   & ($p < 0.01$) \\
\midrule
Scenario 3        & Normal   & $1.48\pm0.14$     & $100.0$ \\
\emph{lose track} & Attack   & $70.9\pm44.4$     & ($p < 0.01$)\\
\bottomrule
\end{tabular}
\end{sc}
\end{small}
\end{center}
\end{table}

\subsection{Running time comparison}
We compared the computation time of generating adversarial images using the iterative optimization-based methods in~\cite{jia2020fooling, jha2020ml} to the generative method in our proposed framework as shown in Table~\ref{tab:runtime}. We deployed the algorithm and ran the simulation environment  on a desktop computer with \emph{AMD Ryzen 5 3600} CPU, 16GB RAM, and \emph{NVIDIA GeForce RTX 2070} GPU. The recursive image attack in Algorithm~\ref{alg:recursive_attack} takes significantly less time per image generation than the other iterative method, as in Table~\ref{tab:runtime}. Due to the long generation time for the iterative optimization method, which exceeds the control loop time of the environment, we first generated the images using the recursive image attack method and applied the iterative optimization method offline. We only used 10 iterations for the iterative optimization.
\vspace{-0.in}
\begin{table}[h]
\caption{Image generation time comparison in Scenario 1. The iterative optimization used 10 iterations.}
\label{tab:runtime}
\vskip -0.2in
\begin{center}
\begin{small}
\begin{sc}
\begin{tabular}{lccc}
\toprule
                    & Recursive     & Iterative \\
                    &Image Attack  & Optimization \\
                    & (avg. $\pm$std.) & (avg. $\pm$std.) \\
\midrule
Run time [$ms$]     & $7.0\pm 0.6$    & $307.5\pm6.3$ \\
Image attack loss   & $17.2\pm4.5$     & $19.5\pm7.2$ \\
\bottomrule
\end{tabular}
\end{sc}
\end{small}
\end{center}
\vskip -0.2in
\end{table}

\section{Conclusion}
In this paper, we proposed a multi-level image attacker consisting of a low-level adversarial image attack generator, an image-based state estimator, and a high-level reinforcement learning attacker. The feed-forward generator can produce adversarial perturbations efficiently compared to the other iterative optimization methods (as claimed in~\cite{xiao2018generating}). Also, the integration of the image-based state estimation and reinforcement learning as the high-level attacker makes our framework is more suitable in situations where state information is less accessible.

\bibliography{mybib}
\bibliographystyle{icml2020}

\end{document}